# Introducing a hybrid model of DEA and data mining in evaluating efficiency. Case study: Bank Branches

Sara Hosseinzadeh Kassani[1]*, peyman Hoseseinzadeh Kassani [2],
seyed Esmaeel Najafi [3]

## Abstract

In today's economy, the banking industry is very important for economic cycle of each country and provides some quality of services for us. With the advancement in technology and rapidly increasing of complexity of today's business environment, it has become more competitive than the past so that efficiency analysis in the banking industry attracts much attention in recent years. From many aspects, such analyses at the branch level are more desirable. Evaluating the branch performance with the purpose of eliminating deficiency can be a crucial issue for branch managers to measure branch efficiency. This work not only can lead to better understanding of bank branch performance, but also give further information to enhance managerial decisions to recognize problematic areas. To achieve this purpose, this study presents an integrated approach based on Data Envelopment Analysis (DEA), Clustering algorithms and Polynomial Pattern Classifier for constructing a classifier to identify class of bank branches. First, the efficiency estimates of individual branches are evaluated by using the DEA approach. Next, when the range and number of classes were identified by experts, the number of clusters is identified by agglomerative hierarchical clustering algorithm based on some statistical methods. Next, we divide our raw data into k clusters By means of self-organizing map (SOM) neural networks. Finally, all clusters are fed into reduced multivariate polynomial model to predict the classes of data.

**Keywords:** Banking, Efficiency, Data Envelopment Analysis, Data Mining, Classification

## Introduction

Banks play a very important role in the economic life of the nations. A bank as a matter of fact becomes very effective partner in the process of economic development of many countries. By technology advancements and the importance of competitive market, banks are able to do investing, lending, borrowing and many functions in a very efficient systematic manner. However, in today's world in which many countries face financial crises, the importance of evaluating the banks' Performance in order to improve their functions and monitor their financial conditions is very important. According to this, evaluating the branch performance with the purpose of eliminating deficiency can be a crucial issue for branch managers to measure branches efficiency.

---

[1] *Department of Management, Electronic Branch, Islamic Azad university, Tehran, Iran
Corresponding Author: Sara Hosseinzadeh Kassani :hosseinzadeh.s@gmail.com  Tel : +98 1333547195

[2] Department of Electrical and Electronics Engineering, Yonsei University, Seoul, Korea
Hosseinzadeh.peyman@gmail.com

[3] Department of Engineering, science and Research Branch, Islamic Azad University, Tehran, Iran
Najafi1515@yahoo.com

Furthermore, evaluating the efficiency in the banking industry has been a focus of numerous research studies and among the long list of modeling techniques in the banking sector, Data Envelopment Analysis (DEA) is one of the most successfully used operational research technique in assessing bank performance (Fethi and Pasiouras, 2010). DEA is one of the methods which is used extensively to evaluate banking and branch efficiency. A lot of papers have published on banking efficiency using the DEA method (Tyrone et al., 2009, Joseph C.Paradi et al., 2012 , Napapan Meepadung, 2009 , Portela and Thanassoulis, 2010 , Deville, 2009).

One of the best ways to improve accuracy is using modular approaches. In modular models, a task or problem is decomposed into a number of subtasks and each module handles a subtask of the global one(Auda and Kamel, 1999). There are different motivations for using modular approaches which are as: to improve performance, to reduce model complexity to make the problem simple to recombine sensory information(Sharkey, 1999). So, in this paper, after modularization of data by using clustering technique, we apply the reduced polynomial model on each subtask (cluster) and merge results to report final accuracy and finally compare the results with non-modular classifier. (Toh et al., 2004)

The rationale why decomposing a big task to some subtasks by clustering technique may improve accuracy is data clustering can divide data into subpopulations and reduce the complexity of the whole data to something more homogeneous(Alpaydin, 2004).

**2. DEA-Clustering-g Framework (Modular DEA)**

This paper presents a modular classifier that we call modular DEA model to construct an expert system for branches classification. There are three main stages to construct this model .The first stage is efficiency measurement stage in which efficiency of banks in Iranian banking industry is measured using data envelopment analysis (DEA). The second stage is data clustering stage where using the knowledge of bank management, bank branches are divided into three classes based on their efficiency scores to convert unsupervised problem to a supervised problem. Now, we use hierarchical clustering algorithm (HCA) to find best number of clusters.

Using self-organization map (SOM) neural network, we divide data into subpopulations and reduce the complexity of the whole data to be more homogeneous. Finally, in classification phase, to prepare data for classification, we try to determine the class label of branches using reduced polynomial classifier. To meet this purpose, each SOM outcome will be individually fed into the classifier.

**2- 1 Efficiency Measurement by DEA**

Data Envelopment Analysis (DEA) was originally proposed by Farell (Farell, 1957) and popularized by Charnes et al. (CCR model) (Charnes et al., 1978)  and Banker et al. (BCC model) (Banker et al., 1984).

DEA is a non-parametric technique that can be used for performance analysis and measuring the relative efficiency of a set of similar units, usually referred to DMUs which convert multiple inputs to multiple outputs. Evaluating DMUs without using any predefined function such as production function is the main advantage of using DEA.

In our study, Estimating efficiency is equal to identify the efficient frontier as a benchmark for measuring relative Performance of the bank branches. The relative efficiency score of a bank branch is determined by how close it is to the efficient frontier.

For measuring efficiency we used an input oriented CCR model. The input-oriented efficiency measure is estimated using the following CCR model (Charnes et al., 1978) of DEA:

$$\theta_j = \text{Min } \theta$$
$$S.t.$$
$$\sum_{j=1}^{n} \lambda_j x_{ij} \leq \theta x_{ij} \quad , \quad i = 1, 2, \ldots, m \tag{1}$$

$$\sum_{j=1}^{n} \lambda_j y_{rj} \geq y_{rj} \quad , \quad r = 1, 2, \ldots, s$$
$$\lambda_j \geq 0 \quad j = 1, 2, \ldots, n$$

Where j is the bank being evaluated, n is the number of DMUs, each DMU consumes m inputs for producing s outputs, $\theta_j$ is the estimated efficiency for bank j, $X_{ij}$ is input i for bank j, $Y_{rj}$ is output r for bank j, and $\lambda$ is the weight placed on banks.

The DMUs with efficiency values equal to 1 are efficient branches, so all DMUs with efficiency values of 1 constitutes a reference set and can be deemed as performance leaders. If a DMU's efficiency value is smaller than 1, it indicates this DMU appears inefficient as compared with the reference set and its efficiency can be improved by means of various management policies.

### 2-2 Set up the Class of Branches by Knowledge of Bank Managers

Branches classification is an effective way for managerial and organizational performance that can be used in budget evaluation, reward system and can provide facilities to monitor operation conditions and financial performance.

After applying DEA and obtaining the efficiency of each branch, these branches are divided into some groups based on their values and consulting with bank experts. Results will be discussed in experiments section.

### 2-3. Clustering Phase
### 2-3-1. The agglomerative hierarchical clustering

Hierarchical clustering analysis (HCA) is a widely used clustering technique for finding the underlying structure of objects through an iterative process. HCA can be divided into two categories: agglomerative methods (AHCA), which make more general groups from all objects (training pattern), and divisive methods (DHCA), which assume all objects as a group and then separate them successively into finer groups. The final output of HCA is a dendrogram or tree diagram. Each level of this dendrogram represents a partitioning of the data set into a certain number of clusters. Therefore, based on the dendrogram it is possible to define the number of clusters. The agglomerative hierarchical clustering algorithm is used in this research to determine the number of clusters (Han and Kamber, 2006).

### 2-4. Classification Phase

As mentioned before, modeling the efficiency of branches for determining the class of branches is a classification problem from data mining viewpoint.

In this stage a Reduced Multivariate Polynomial Pattern Classifier is used to model the class of the branches. In following, we'll describe multivariate polynomial model and its reduced version in a nutshell.

#### 2-4-1. Multivariate Polynomial Model

The Multivariate Polynomial model (MP) provides an effective way to describe complex nonlinear input-output relationships. However, for high-dimensional and high-order systems, multivariate polynomial regression becomes impractical due to its huge number of product terms.

The general multivariate polynomial model can be expressed as

$$g(\alpha, x) = \sum_{i}^{K} \alpha_i x_1^{n_1} x_2^{n_2} \ldots x_l^{n_l} \tag{2}$$

Where the summation is taken over all nonnegative integers $n_1, n_2 \ldots, n_l$ for which $n_1 + n_2 + \ldots + n_l \leq r$ with r being the order of approximation. $\alpha = [\alpha_1, \ldots, \alpha_k]^T$ is the parameter vector to be estimated and x denotes the regressor vector $[x_1, \ldots, x_l]^T$ containing l inputs. K is the total number of terms in $g(\alpha, x)$. Consider a polynomial model given by

$$g(\alpha, x) = \alpha^T \, p(x) \tag{3}$$

Where alpha is coefficients and p(x) is regression vector. The goal is estimating coefficients.

Given m data points and using the least-squares error minimization objective given by

$$S[\alpha, x] = \sum_{i=1}^{m} [y_i - g(\alpha, x_i)]^2 = [y - P\alpha]^T [y - P\alpha], \quad (4)$$

The parameter vector α can be estimated from:

$$\alpha = (P^T P)^{-1} P^T y, \quad (5)$$

Where $P \in R^{m \times K}$ denotes the Jacobian matrix of $P(x)$.
And $y = [y_1, \ldots, y_m]^T$ is the known inference vector from training dara.

It is noted here that (5) involves computation of the inverse of a matrix, the problem of multicollinearity may arise if some linear dependence among the elements of $x$ are present. A simple approach to improve numerical stability is to perform a weight decay regularization using the following error objective:

$$S[\alpha, x] = \sum_{i=1}^{m} [y_i - g(m, x_i)]^2 + b \|\alpha\|_2^2$$
$$= [y - P\alpha]^T [y - P\alpha] + b\alpha^T \alpha, \quad (6)$$

Where $\|.\|_2$ denotes the $l_2$-norm and b is a regularization constant.
Minimizing the new objective function (8) results in

$$\alpha = (P^T P + bI)^{-1} P^T y, \quad (7)$$

Where $P \in \mathcal{R}^{m \times K}, y \in \mathcal{R}^{m \times 1}$ and $I$ is a $(K \times K)$ identity matrix. This addition of a bias term into the least-squares regression model is also termed as *ridge regression*.

### 2-4-2. Reduced Multivariate polynomial Model

For a rth-order model with input dimension $l$, the number of independent adjustable parameters would grow like $l^r$ (Bishop, 1995). For medium to large sizes of data dimensions, the MP model would need a huge quantity of training data to ensure that the parameters are well determined. In view of this problem, we resort to possible reduced models whose number of parameters do not increase exponentially and, yet, preserving the necessary classification capability.

To resolve this problem, a linearized model is considered.

Given two points α and $\alpha_1$ on the multinomial function which is differentiable. By the Mean Value Theorem, the multinomial function $f(\alpha) = (\alpha_{j1} x_1 + \alpha_{j2} x_2 + \ldots + \alpha_{jl} x_l)^j$, $j = 2, \ldots, r$ (indicating only the regressor parameter to simplify the expression) about the point $\alpha_1$ can be written as:

$$f(\alpha) = f(\alpha_1) + (\alpha - \alpha_1)^T \nabla f(\bar{\alpha}), \quad (9)$$

Where $\bar{\alpha} = (1 - \beta)\alpha_1 + \beta\alpha$ for $0 \leq \beta \leq 1$. Let $x = [x_1, \ldots, x_l]^T$. By omitting the reference point $\alpha_1$ and those coefficients within $f(\alpha_1)$ and $\nabla f(\bar{\alpha})$ and including the summation of weighted input terms, the following multivariate model can be written:

$$\hat{f}_{RM'}(\alpha, x) = \alpha_0 + \sum_{j=1}^{l} \alpha_j x_j + \sum_{j=1}^{r} \alpha_{l+j}(x_1 + x_2 + ... + x_l)^j$$
$$+ \sum_{j=2}^{r}(\alpha_j^T . x)\left((x_1 + x_2 + ... + x_l)^{j-1}\right), l, r \geq 2, \quad (9)$$

Where the number of terms is given by $k = 1 + r(l+1)$. To include more individual high-order terms for (9), the following (RM) can be written:

$$\hat{f}_{RM}(\alpha, x) = \alpha_0 + \sum_{k=1}^{r}\sum_{j=1}^{l} \alpha_{kj} x_j^k + \sum_{j=1}^{r} \alpha_{rl+j}(x_1 + x_2 + ... + x_l)^j$$
$$+ \sum_{j=2}^{r}(\alpha_j^T . x)\left((x_1 + x_2 + ... + x_l)^{j-1}\right), l, r \geq 2, \quad (10)$$

The number of terms in this model can be expressed as: $k = 1 + r + l(2r - 1)$. It is noted that (10) has $(rl - l)$ number of terms more than that of (9). Now, the reduced model is much faster than traditional one and is less negatively affected by ill-conditioning and overfitting problem since it uses much less parameters for tuning.

### 3- Experimental Results
### 3.1. Data collection

The collaborating Bank in this study operates 589 branches located in Iran and offers a wide variety of services including investing, lending, cash management, deposits and etc.

List of input variables and their definitions is given in Table1.

**Table1. The Summary of Dataset Variables**

| Variable Name | Definition |
|---|---|
| Interest rate $(I_1)$ | It is a ratio that indicates the rate of customers' deposits. |
| Nonoperational cost for earning financial resources $(I_2)$ | Non-operational expense of resources absorption or customers' deposit. |
| Staff Cost $(I_3)$ | The input costs and additional salary and bonuses paid to the personnel division. |
| Rate of Islamic contracts $(O_1)$ | Interest rate in the contracts between the customer and the bank that will be received from customer. |
| Growth of financial resources $(O_2)$ | Refer to below of this table. |
| Growth of loans $(O_3)$ | Refer to below of this table. |

Referring to above table, I's show the inputs of DEA and O's show its outputs. In general, improvement of branch performance by maintaining a consistent or constant trend is very valuable.

### 3.2 The role of DEA and data modularization

As before mentioned, for measuring efficiency we used an input oriented CCR model. After applying this model on branches data, 10 branches were identified 100% efficient.

After consulting with bank management and the benefit of their knowledge, branches were divided into three groups. The ranges of efficiency scores were: $[0,0.55)$, $[0.55,0.7)$, $[0.7,1]$, average and high performance respectively. Now, our unsupervised problem was turned to supervised problem which is addressed to classification.

In data preprocessing stage, we use clustering algorithms to obtain the number of clusters and the cluster label of samples. Then, we use each cluster to do classification task to develop a modular classifier. We use AHCA algorithm to obtain suitable number of clusters based on some important statistical analysis. Applying some statistical cluster analyses, results show three number of clusters are suitable for doing clustering task by SOM.

Table 2 shows the clusters members and R-square values for each cluster. Three indicators are available for this type of analysis: $R^2$ with own cluster, $R^2$ with the nearest cluster and the $1 - R^2$ ratio. The first indicator shows the variables of each cluster how much correlated with own cluster. Second shows the variables of each cluster how much correlated with nearest cluster. This column also has a small value for each variable that is good again. Small value of third indicator indicates good clustering. If this value is larger to 1, it means that the variable has a larger correlation with another cluster than its group. All indicators indicate we haven't clustering heterogeneity.

**Table2. Cluster members and R-square values**

| Cluster | Members | Own cluster | Next closest | 1-$R^2$ ratio |
|---|---|---|---|---|
| 1 | L3 | 0.501 | 0.0088 | 0.4635 |
| 1 | O2 | 0.501 | 0.0630 | 0.4926 |
| 2 | L1 | 0.614 | 0.0810 | 0.4204 |
| 2 | O1 | 0.614 | 0.0060 | 0.3887 |
| ٣ | L2 | 0.606 | 0.0671 | 0.4215 |
| ٣ | O3 | 0.606 | 0.0102 | 0.3973 |

Table 3 shows the cluster correlation for each cluster. It displays the correlation of each variable to clusters. It enables us to interpret the clusters. The correlations higher than 0.7 or lower than -0.7 are in bold face and we count these situations in the Membership column. If the variables are well clustered, each variable must be associated to one and only one cluster.

**Table3. Structure of Cluster correlation:**

| Attribute | # Membership | Cluster 1 | Cluster 2 | Cluster 3 |
|---|---|---|---|---|
| I1 | 1 | -0.0001 | **0.7834** | -0.2846 |
| I2 | 1 | 0.0326 | -0.2590 | **0.7790** |
| I3 | 1 | **0.7078** | -0.1024 | -0.0939 |
| O1 | 1 | -0.1407 | **0.7834** | -0.0777 |
| O2 | 1 | **0.7078** | -0,0248 | 0.2511 |
| O3 | 1 | 0.1403 | -0.1012 | **0.7790** |

In our dataset, we obtained a good association between variables and their cluster and all variables are highly correlated to their clusters.

After identifying the number of clusters according to the AHCA, we perform SOM neural network clustering. Learning rate of 0.03 was assumed. After clustering, the first cluster has given 227 and the second cluster has 241 and the third cluster has 121 samples.

However, to prevent overpowering, the data is often normalized. Normalization of data was based on the distance and a variance-based normalization was performed.

After using SOM neural network and cluster the branches to 3 different groups, the results are shown in Table4.

Table4. Distribution of branches in each cluster

| Number of Branches | Cluster 1 | Cluster 2 | Cluster 3 |
|---|---|---|---|
| weak performance | 22 | 10 | 7 |
| average performance | 188 | 199 | 80 |
| high performance | 17 | 32 | 34 |

Now we use RM model on each cluster and all data to compare modular and non-modular methods.

### 3.3. Developing modular classifier using the proposed model

To demonstrate the feasibility and effectiveness of the proposed model, we used 10-fold cross validation for training classifiers for all clusters. For reporting overall accuracy, we use weighted averaging method. Indeed, we average the results after getting all these k=10 results for each class based on the weight of that class. The weighted averaging cross validation error rate is obtained based on equation:

$$\text{Weighted averaging} - \text{K} - \text{CV error} = \frac{1}{K}\sum_{k=1}^{K} W_{kj} \cdot \text{error}(k) \tag{11}$$

$$j = 0,1$$

Where $\text{error}(k)$ is the error rate of $k^{th}$ testing set and $W_{kj}$ is the proportion of samples belong to one of these classes in the $k^{th}$ fold and j is the name of classes. The main advantage is preserving the reliability of the estimation.

In order to investigate the effects of modular classifier on improvement of the proposed model, we compare modular and non-modular models. We use weighted average for three clusters as performance measure of this modular model. Classification accuracy (CA) measurement for a modular model is obtained as follows:

$$\begin{aligned}CA\ of\ modular\ mode = &\ W_1 \times CA(cluster\ 1) + \\ &\ W_2 \times CA(cluster\ 2) + \\ &\ W_3 \times CA(cluster\ 3)\end{aligned} \tag{12}$$

Where:

$$W_1 = \frac{Number\ of\ branches\ in\ cluster\ 1}{Total\ Number\ of\ branches}$$

$$W_2 = \frac{Number\ of\ branches\ in\ cluster\ 2}{Total\ Number\ of\ branches} \tag{13}$$

$$W_3 = \frac{Number\ of\ branches\ in\ cluster\ 3}{Total\ Number\ of\ branches}$$

First, the classification accuracy for these two approaches is reported in Table5.

Table5. The results of classification accuracy

| Model | Non-Modular | Cluster1 | Cluster2 | Cluster3 |
|---|---|---|---|---|
| Correctly Classified Instances | 527 (85.75 %) | 212 (90.27 %) | 227(89.45 %) | 111 (88.67 %) |

All results are in polynomial order 2 which is the best order obtained in terms of classification accuracy. For the sake of comparison using eq. 12, we can report only one value for modular based model as follows:

$$(212/527)*0.9027 + (227/527)*0.8945 + (111/527)*0.8867 = 0.9352$$

Relying on CA, the modular model (0.9352) is very satisfactory and outperforms non modular one (0.8575).

### 4- Conclusions

This study provides a new model for measuring efficiency. It compares two approaches for measuring efficiency: modular and non-modular. To construct models, first, the concept of efficiency was measured by DEA. Next, two clustering algorithms were used. We used HCA for identifying suitable number of clusters and SOM for doing clustering task.

In classification section, we used RM model to get classification accuracy to investigate modular and non-modular approaches. The experimental results demonstrate that the modular approach performs significantly better than the non-modular approach in terms of classification accuracy. The most important reason for this achievement is our clustering task that was very successful and it could produce homogenous clusters.

**References:**

ALPAYDIN, E. 2004. Introduction to machine learning MIT Press.

AUDA, G. & KAMEL, M. 1999. Modular neural networks: A survey. *International Journal of Neural Systems,* 9**,** 129–151.

BANKER, R. D., CHARNES, A. & COOPER, W. W. 1984. Some Methods for Estimating Technical and Scale Inefficiencies in Data Envelopment Analysis. *Management Science,,* 30**,** 1078-1092.

BISHOP, C. M. 1995. Neural networks for pattern recognition.

CHARNES, A., COOPER, W. W. & RHODES, E. 1978. Measuring the effciency of decision making units,. *European Journal of Operational Research,* 2**,** 429–444.

DEVILLE, A. 2009. Branch banking network assessment using DEA: A benchmarking analysis - A note *Management Accounting Research,* 20**,** 252–261.

FARELL, M. 1957. The measurement of productive effciency,. *Journal of the Royal Statistical Society, Series A (General),* 120**,** 253–281.

FETHI, M. & PASIOURAS, F. 2010. Assessing bank efficiency and performance with operational research and artificial intelligence techniques: a survey. *European Journal of Operational Research,* 204**,** 189–198.

HAN, J. W. & KAMBER, M. 2006. *Data mining: Concepts and techniques,*.

JOSEPH C.PARADI, HAIYAN ZHU & EDELSTEIN, B. 2012 Identifying managerial groups in a large Canadian bank branch network with a DEA approach. *European Journal of Operational Research,* 219 178-187.

NAPAPAN MEEPADUNG, J. C. S. T., DOBA KHANG 2009 IT-based banking services: Evaluating operating and profit efficiency at bank branches. *Journal of High Technology Management Research,* 20**,** 145–152.

PORTELA, M. C. A. S. & THANASSOULIS, E. 2010 Malmquist-type indices in the presence of negative data: An application to bank branches. *Journal of Banking & Finance,* 34**,** 1472–1483.

SHARKEY, A. J. C. 1999. *Combining artificial neural networks*, Springer.

TOH, K.-A., TRAN, Q.-L. & SRINIVASAN, D. 2004. Benchmarking a reduced multivariate polynomial pattern classifier. *Pattern Analysis and Machine Intelligence, IEEE Transactions on,* 26**,** 740-755.

TYRONE, L., CHIA-CHI, L. & TSUI-FEN, C. 2009. Application of DEA in analyzing a bank's operating performance. *Expert Systems with Applications,* 36**,** 8883–8891.